\documentclass[]{iscram}

\usepackage[utf8]{inputenc}

\usepackage{lipsum}
\usepackage{tikz}
\usepackage{fancyvrb}
\usepackage{listings}
\usepackage{enumitem}
\usepackage{tcolorbox}
\usepackage{multicol}
\usepackage{siunitx}

\usepackage{graphicx}

\lstdefinestyle{common}{
  xleftmargin=.5em,
  xrightmargin=.5em,
  frame=single,framesep=.5em,framerule=0pt,
  fancyvrb=true,
  basicstyle=\ttfamily,
  keywordstyle=\color{cyan!50!blue!75!black}\bfseries,
  commentstyle=\color{red!50!black}\itshape,
  stringstyle=\ttfamily\color{green!50!black},
  numbers=none,
  showspaces=false,
  showstringspaces=false,
  fontadjust=true,
  keepspaces=true,
  flexiblecolumns=true,
  emphstyle=\color{red},
}
\lstdefinestyle{TeX}{
  style=common,
  backgroundcolor=\color{blue!5},
  aboveskip=5pt,
  belowskip=5pt,
  language=[LaTeX]TeX,
  moretexcs={
    abstract, addbibresource, iscramset, keywords, mainmatter,
    maketitle, printbibliography, subsection, subsubsection, url,
    urldef, href, includegraphics, ldots, parencite, citeauthor,
    citeyear, citetitle, midrule, toprule, bottomrule
  },
  fancyvrb=true,
}
\lstdefinestyle{console}{
  style=common,
  backgroundcolor=\color{gray!10},
  aboveskip=5pt,
  belowskip=5pt,
}

\newlist{options}{description}{1}
\setlist[options]{%
  beginpenalty=10000,%
  itemsep=.5\parskip plus .3\parskip minus .2\parskip,
  parsep=.5\parskip plus .3\parskip minus .2\parskip,
  topsep=.5\parskip plus .3\parskip minus .2\parskip,
  partopsep=.5\parskip plus .3\parskip minus .2\parskip,
  style=nextline,labelindent=1em,%
  font=\normalfont\ttfamily}

\colorlet{macro color}{cyan!50!blue!75!black}
\colorlet{option color}{red!50!black}
\colorlet{generic color}{green!40!black}

\newtcolorbox{pseudoTeX}{colback=blue!5,colframe=blue!5,before=\nobreak}
\let\LaTeXorig\LaTeX
\renewcommand\LaTeX{\bgroup\fontfamily{lmr}\selectfont\upshape\LaTeXorig\egroup}

\urldef{\sitewebgind}\url{http://gind.mines-albi.fr}
\urldef{\sitewebminesalbi}\url{http://www.mines-albi.fr}
\urldef{\gmailpaulgaborit}\url{paul.gaborit@gmail.com}
\urldef{\jdmail}\url{j.doe@example.com}
\urldef{\sitex}\url{www.example.com}

\iscramset{
  title={
    Helping Crisis Responders Find\\the Informative Needle\\in the Tweet Haystack
  },
  short title={Finding Informative Tweets},
  author={
    short name={Leon Derczynski},
    full name=Leon Derczynski\thanks{corresponding author},
    affiliation={University of Sheffield},
  },
  author={
    full name=Kenny Meesters,
    affiliation={TU Delft},
  },
  author={
    full name=Kalina Bontcheva,
    affiliation={University of Sheffield},
  },
  author={
    full name=Diana Maynard,
    affiliation={University of Sheffield},
  },
  WiPe Paper 2018=Social Media Studies 
}

\usepackage{tikzpagenodes}
\usetikzlibrary{fit}

\begin{document}

\maketitle

\makeatletter
{\centering\large\iscram@version{}\\\iscram@date\par}
\makeatother

\abstract{
Crisis responders are increasingly using social media, data and other digital sources of information to build a situational understanding of a crisis situation in order to design an effective response. However with the increased availability of such data, the challenge of identifying relevant information from it also increases. This paper presents a successful automatic approach to handling this problem. Messages are filtered for informativeness based on a definition of the concept drawn from prior research and crisis response experts. Informative messages are tagged for actionable data -- for example, people in need, threats to rescue efforts, changes in environment, and so on. In all, eight categories of actionability are identified. The two components -- informativeness and actionability classification -- are packaged together as an openly-available tool called Emina (Emergent Informativeness and Actionability).
}

\keywords{informativeness, twitter, social media, actionability, information filtering}

\section{Introduction}
A key aspect of dealing with the aftermath of disasters and critical events is to get a quick situational understanding of the situation. This situational understanding enables responders to not only understand the gravity of the situation and the needs of the affected community, but also the operational circumstances and the availability of resources. However, during a crisis situation getting a coherent understanding of the situation can prove challenging, due to its chaotic nature. Therefore crisis responders are continuously looking for key pieces of information to build, implement and improve their response~(\cite{harrald2006}).
 
In recent years, the advent of information technology has provided crisis responders with new tools and systems to collect information directly from the communities affected by crisis. At the same time, with the rise of social media and the profusion of mobile technology the same communities are also increasingly empowered -- and at times incentivized -- to share their data through an increased number of channels. These developments provide crisis responders with access to more data than ever before, directly from the communities affected by the crisis~(\cite{vieweg2010}).

However, this data surge has given rise to new challenges. Where before the key challenges were to gather sufficient data to build a situational understanding, the main challenges now are to identify the relevant messages~(\cite{comes2017}). While many responding agencies have confirmed the usefulness of social media and other forms of electronic information gathering, it remains challenging to integrate an effective workflow in short-burning crisis situations~(\cite{meesters2016}). Currently, organizations often rely on external volunteers, such as the Standby Task Force in the wake of the 2010 Haiti Earthquake~(\cite{patrick2011,rencoret2010}). However, not all organizations have the capacities or knowledge to process the large amounts of data available to them in the search for relevant messages. 

Information filtering is critical to crisis workers dealing with streams of information related to emerging and ongoing situations~(\cite{vieweg2012, temnikova2015}). Today, messages in many forms arrive constantly in volumes too large for humans to effectively prioritize in real-time. In this research, we aim to provide automated methods for filtering this information so that crisis workers can effectively find potentially important messages and perform information triage.

Crisis worker flow typically begins with defining selectors that determine what information is retrieved; these could be keywords, particular sources, GPS coordinates, and so on. For example, to get content about hurricane Irma, one could capture just messages that contain ``irma" and originate within the monitored area. This is the first pass at reducing the volume of information, but even with this, the volume of information balloons rapidly as events progress; data for Typhoon Yolanda in 2013 goes from 182 tweets in the first three days to 1778 on the fifth day and 4385 a week later; the west Texas explosion generated 10K tweets on the first day, going from 128 tweets in hour 1 to 2456 in hour 4.\footnote{From CrisisLex data~(\cite{olteanu2014})} And indeed these selectors do not guarantee that the incoming data is good; many messages can be irrelevant~(\cite{vandenhomberg2014}). 

These challenges are however not only related to natural disasters, but also extend to other disruptions and incidents faced by communities around the world. For example, in a recent iHub project exercise monitoring elections in Kenya, workers found roughly 90\% of the messages captured in the message monitoring Ushahidi platform were irrelevant\footnote{See https://www.ushahidi.com/blog/2017/08/10/ushahidis-global-uchaguzi-partnership-crowdsourcing-kenyan-election-incidents-shows-a-mostly-orderly-election} -- not only due to their content, but also due to missing content, repetitive messages, and so on.  In this case the messages were deliberately solicited: people were encouraged to post their own observations (via various channels). Despite this 'top-down' approach, the collected data still resulted in the high percentage of irrelevant messages. In sudden onset disasters, the number of messages and noise is potentially much higher. Methods to ease the impact of high data volume are therefore still pressing. 

These developments and trends show that a tool to reduce the amount of data seen by humans and filter to just the relevant messages is not only necessary  but would improve the usefulness of the data available to responders. The development of auto-assessment of informativeness is the first challenge that needs to be addressed towards a system of automated identification of relevant messages for crisis responders.

In this paper, we describe the openly-available Emina (Emergent Informativeness and Actionability) tool, which we have developed to address these issues as part of the EU COMRADES project.\footnote{See http://comrades-project.eu}

\section{Development Approach}
Our approach began by developing a definition for informativeness. With this, we can gather and annotate data for its informativeness, using a crowdsourcing task, where readily-available workers are drawn from a pool over the internet to quickly process the kinds of tasks that require human intelligence. This produces our training set. This training set is used to train a machine learning system, which learns mathematical rules about the text in the training set that help it predict the informativeness of future, previously-unseen messages, mimicking a deployed situation. The same general approach is taken to learning actionability.

The developed method for assessing the informativeness  and filtering the messages is divided into two stages. Firstly, incoming messages are labelled for informativeness by a machine learning system. Messages deemed to be uninformative are discarded. The sensitivity of this system can be tuned. For example, in situations where it is critical to process every potentially important message, this can be achieved, at the cost of some irrelevant messages coming in. In other situations, where it’s most important to keep up to date, and only summary information is required, then all irrelevant messages can be discarded at the risk of some potentially relevant ones being missed.

The second stage is to determine what kind of action can be taken based on the message. This is termed {\em actionability classification}. This paper describes a novel schema for actionability classification. The schema is prototyped, updated, and then used to construct a dataset. A machine learning system is then trained to classify incoming messages according to their actionability, based on this dataset.

\section{Predicting Informativeness}

Discarding uninformative data is critical when one aims to process crisis-relevant messages in real time. Typically, when capturing messages related to a crisis, any potential selector (e.g. geographical area, hashtag) yields a high volume of available content, via e.g. Tweets or SMS. However, only a limited part of this will in fact be relevant to the crisis and also informative.

Therefore, the first step in dealing with any information stream related to a crisis is to differentiate between informative and non-informative content. The two major technical challenges here are first in developing a computationally operationalised definition of informativeness in crisis situations, and secondly in creating a technical solution that can discriminate based on informativeness in the extraordinarily broad range of not only crisis situations but also the social and other message media used around them.

\subsection{Defining Informativeness}
A key challenge in developing an automated approach for identifying relevant messages is to describe ‘relevant’ in a manner that enables it to be transformed into an automated process. However, identifying messages which are relevant, or informative (having an information value), is often both heavily context-dependent and also subjective. Furthermore, informativeness patterns (and associated relevancy of information) may differ from one type of crisis to another~(\cite{olteanu2015}). Because we rely partially on human-annotated data to train our system, the definition of informativeness must additionally both provide useful distinctions for the end users of the tool (the crisis workers) and be as easy as possible for human annotators to understand and apply to the data they are given. Previous work has shown that the quality of crowdsourced data is heavily dependent on both the simplicity and clarity of the task and definition~(\cite{sabou2014}), and this data is critical for the development of the tool.

\begin{figure}
\centering
\includegraphics[width=0.6\columnwidth]{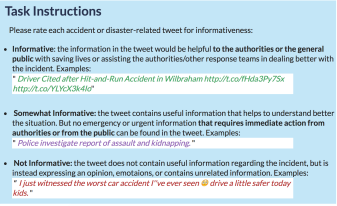}
\caption{How the informativeness task is presented to crowd workers}
\label{fig:inf}
\end{figure}

These issues makes it challenging both to capture good examples of informativeness, and to develop a definition for it that can be used for crisis analysis. For this scenario, the notion of informativeness ideally needs to reflect the extent to which information captured during a crisis is useful to responders; or at least whether or not a message is related to the crisis. Rather than build a definition from first principles, we instead took a top-down approach, considering what kinds of messages in practice would be useful, and generalising from that. To arrive at a definition, we drew on a range of existing sources and efforts. These included third-party resources from established authorities in the field of information management for crisis response, such as CrisisLex~(\cite{olteanu2014}), the OCHA Guiding Principles~(\cite{vandewalle2008}), and the World Disaster report, as well as academic sources such as the DERMIS principles~(\cite{turoff2004}), and field research in Nepal~(\cite{baharmand2016}). Finally, we included research carried out within the COMRADES project, including monitoring exercises from iHub and input from Delft University of Technology~(\cite{comes2016}).

From these resources, and after some trial and error testing different definitions on a number of volunteer annotators and getting feedback, we created a three-level definition of informativeness. It became clear that informativeness often hinges on how urgent information is, so this concept is reflected in the levels:

\begin{itemize}
\item {\bf Informative:} the information in the message would be helpful with saving lives or assisting the authorities/other response teams in dealing better with the incident.
\item {\bf Somewhat Informative:} the message contains useful information that helps to understand better the situation. But no emergency or urgent information can be found in the tweet.
\item {\bf Not Informative:} the message does not contain useful information regarding the incident, but is instead expressing an opinion or contains unrelated information
\end{itemize}

Because we crowdsource the work to human annotators, we were mindful of how task design and framing could affect results: taking care of cognitive biases is paramount to achieving good results~(\cite{sabou2014, derczynski2016, eickhoff2018}). Specifically, we wanted annotators to think carefully about mid-cases. Our initial tests showed that a binary decision (informative or not) was too rigid; on the other hand, giving annotators too many categories made it hard to find the boundaries between them. A sample task is shown in Figure~\ref{fig:inf}.

\subsection{Data Collection}
The text data used in this stage of the project was part of the dataset released in~\cite{schulz2017}. This dataset includes over 10K tweets that were selected and labelled for the domain of incident detection of 10 different cities. The tweets were already labelled by crowdsourced annotators, and classified. The first dataset has two classes (related and not related to the incident); the second has four classes (crash, fire, shooting, and not related). Note that there is a distinction between related and informative: related messages are not necessarily informative, but unrelated messages are definitely not informative. We thus retrieved 600 tweets from the first dataset taking only those labelled as related. Following this, we filtered our dataset to remove all retweets, as well as tweets with redundant content (i.e. subjectively carrying the same information or describing the same fine-grained event).

Annotators on CrowdFlower\footnote{http://www.crowdflower.com/} were then instructed to label these documents according to the three informativeness categories listed above. Five crowd worker judgments were collected for each tweet. Crowd responses were then adjudicated taking the most popular category. In general, total agreement was higher than 70\% in all three groups, and over 80\% for the ``not informative" category. The resulting annotated data has been made publicly available as the COMRADES Crowdsourced Informativeness Dataset (CCSID)~(\cite{qarout2018}).

\subsection{Supplemental Data}
At least a few thousand examples are required to train most machine learning systems. For extra data, we used CrisisLex~(\cite{olteanu2014}), which contains around 60,000 tweets. These are labelled for informativeness, though not as stringently as CCSID; messages are coded as either ``Informative", corresponding to the union of CCSID’s first two options, or ``Non-informative", corresponding to the third. In any event, it comprises 32K positive and 28K negative examples. While this dataset is not as reliable (for example, there are anecdotal cases of useless ``informative" tweets in it), the effect of this is diminished by balancing data and exploiting loss functions during training. This helps us to get high quality from the CCSID while leveraging the bulk of the CrisisLex annotations, as described below.

\begin{figure}
\centering
\includegraphics[width=0.6\columnwidth]{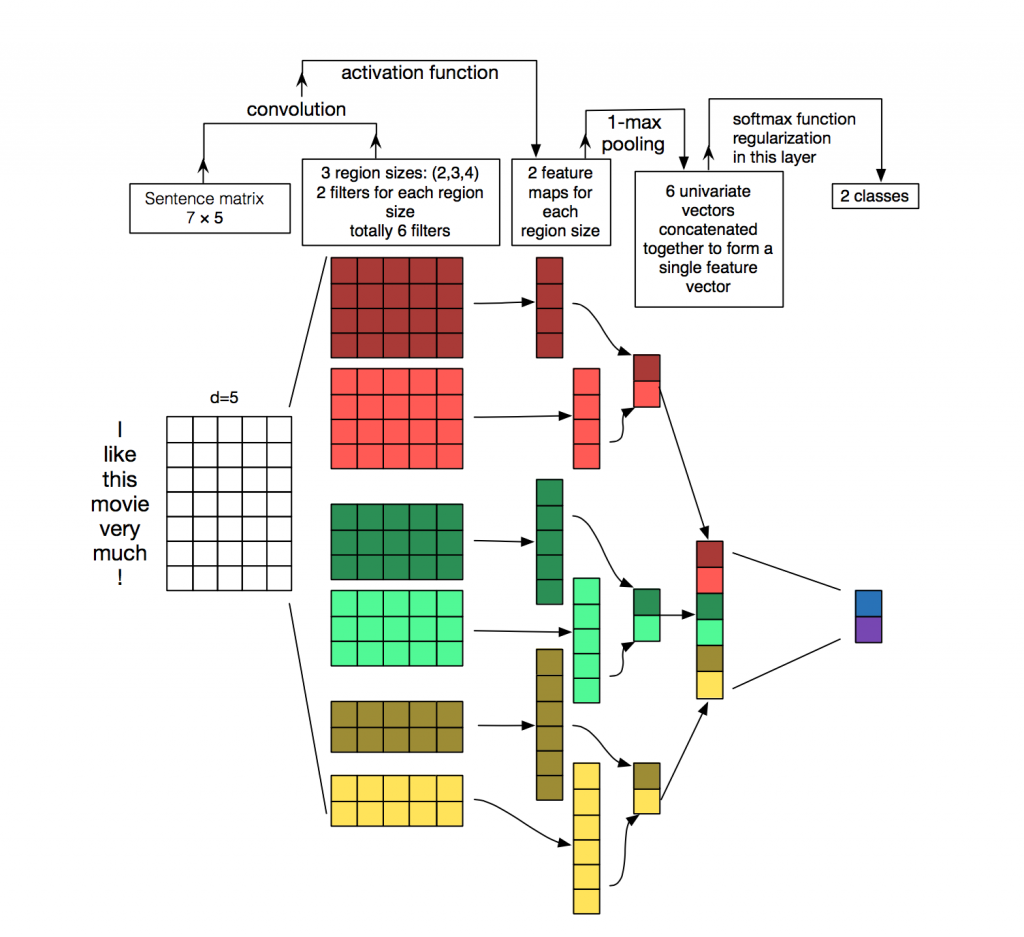}
\caption{The convolutional neural network architecture used (from Zhang et al. 2015)}
\label{fig:cnn}
\end{figure}

\subsection{Machine Learning Approach}
Having the data, the next step is to learn what informative messages generally look like. A machine learning algorithm will pick up patterns in the data we present it with, and try to generalise from that what signals an informative message. Then, when applied to new messages, it should be able to make some decision about what is and is not informative. In this instance, because we are dealing with highly diverse text and have a reasonable amount of data, we choose a convolutional neural network, which ought to be able to infer directly from text without additional feature extraction.

Because we need to leverage the data from CrisisLex as well as CCSID, we can label a maximum of two classes, as CrisisLex does not have the middle ``informative but not urgent" label. Therefore, for this system, we reduce the CCSID categories from three to two by combining ``informative and urgent" with ``informative but not urgent".

Convolutional networks work by repeatedly applying a ``window" to a numerical representation, and combining all the values under that window. For text, those numbers come from a word embedding. We followed the convolutional text classification network of~\cite{zhang2015} to determine informativeness (Figure~\ref{fig:cnn}).

It is important to know how good a machine learning model is, so that we know when to stop training. Sometimes, training for more time does not give a better result. This is because some systems match the training examples too closely, and so will not operate well in a new environment. For example, if our training set only took examples from shootings, and the term ``AK47" occurred often in this data, then a system might end up thinking that highly specific mentions of ``AK47" indicate informativeness, instead of phrases like ``wounded". This is called overfitting.

The quality of a machine learning model is judged in two ways. First, we can measure how well the trained model works on the data that is used to train it (the ``training set"). This is like measuring how well the system has learned the data to which it has been exposed. The metric used for this is ``training loss", and lower is better. The second way is to measure the model’s performance on some other, held-out data (a ``validation set"). A good score on this indicates that the model is learning to predict things beyond the data it has already seen, i.e. the model is generalising. This is called ``validation loss", and again, a lower value is better.

Typically, one takes a dataset, and splits off a part which becomes the validation set. 
During training, training loss decreases, as does validation loss. When  training loss becomes lower than the validation loss, the model has started to predict the training data better than other data, meaning it has started overfitting. So, the model selected is the one just before this crossover of validation and training loss, with a minimum validation loss.

In the instance of our informativeness classifier, we have a somewhat limited dataset (large scale machine learning text datasets typically comprise millions of examples). In addition, some of the training data (ie. from CrisisLex) is not a precise match for the kinds of scenarios we envisage COMRADES outputs being deployed in. This means that measuring loss against CrisisLex alone might gives a system that is very good for the kind of scenarios contained in this dataset, but does not generalise well across crises.

To remedy this, we use part of the CCSID in the validation set, and give it a much larger proportion in the validation set than the test set. Specifically, the validation set is comprised of 300 CCSID instances and 300 drawn from CrisisLex (150 each of informative and non-informative). The remainder of the data, some 60,000 examples, forms the training data. This means that scoring well on the more-varied CCSID, which matched COMRADES goals well, is very important, despite it only comprising a smaller proportion of the training data. In this way, we can select a model for informativeness classification that uses all the data we have available and is optimally likely to score well in real-world deployment.

\subsection{Evaluation}
This training and validation partition were used with the configuration specified above. The final scores in this case are an accuracy of 92\% and an F1 of 91\%. This is over the validation dataset, which represented both the definition of informativeness we wanted to capture and also the diversity present in social media. The F1 indicates that we achieve an excellent balance in errors between false positives (i.e. uninformative messages slipping through) and false negatives (informative messages being missed). We attribute the high performance of this system to the careful annotation process, as well as the volume and range of documents available for training. Additionally, we purposefully created a system aimed at high performance on this data type.

\section{Actionability}
Having filtered out uninformative content, the next stage is to assist crisis workers in information processing by categorizing the remaining messages. This could be used to select individual messages and hand them over to a specialist team, as well as to get a good overview of the crisis situation or detect emerging events.

Some former work has addressed this. Recently, the CREES system~(\cite{burel2017}) used the CrisisLex categories and large amount of data there to do some filtering of information into different categories. However, not all these were actionable, and while there is some overlap, the categories are either not directly useful when determining actionability, or do not provide a sufficient amount of granularity. Nevertheless, with its large amount of data, CREES achieves good performance. Earlier work on determining actionability in crises was effective though studied over a single crisis~(\cite{munro2011}). We go further and train on multiple crises, allowing a more general impression of how actionability can look over a broad range of contexts.

\subsection{Defining Actionability}
Our requirements were that a relatively small number of categories should be selected, and all of these at a single, top level. A hierarchical system is complex and hard to build; the added complexity here incurs cost and quality risk, and for a real-time prototype system, it is better to provide something that works. Further subcategorization could always be efficiently provided later if necessary.

\begin{figure}
\centering
\includegraphics[width=0.95\columnwidth]{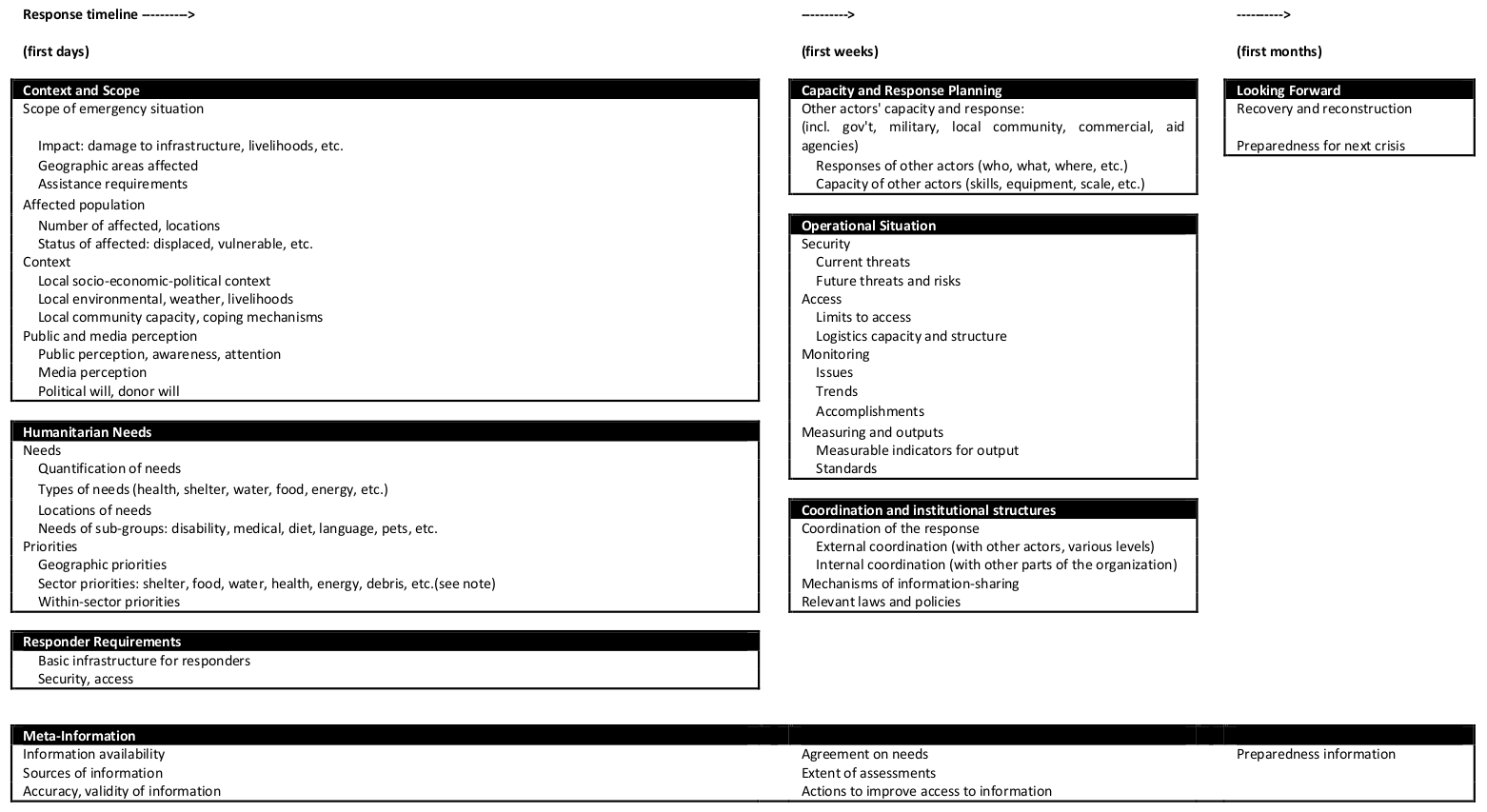}
\caption{A schema from the MIT Humanitarian Response Lab for prioritising hurricane response}
\label{fig:mithr}
\end{figure}

Additionally, the categories should correspond to issues relevant to the type of crisis tracked by COMRADES and envisaged for use by project stakeholders. This means some categories in other datasets feature less in our work. For example, media opinion of the crisis response is less important in our use cases than trapped and injured people.
The source categories are given here; firstly, for the MIT Humanitarian Response Lab’s schema, as in Figure~\ref{fig:mithr}.\footnote{See https://humanitarian.mit.edu/ (MIT HR is also the source of the schema diagram)}

Secondly, from CrisisNLP~(\cite{imran2016}):

\begin{enumerate}
\item Injured or dead people: Reports of casualties and/or injured people due to the crisis
\item Missing, trapped, or found people: Reports and/or questions about missing or found people
\item Displaced people and evacuations: People who have relocated due to the crisis, even for a short time (includes evacuations)
\item Infrastructure and utilities damage: Reports of damaged buildings, roads, bridges, or utilities/services interrupted or restored
\item Donation needs or offers or volunteering services: Reports of urgent needs or donations of shelter and/or supplies such as food, water, clothing, money, medical supplies or blood; and volunteering services
\item Caution and advice: Reports of warnings issued or lifted, guidance and tips
\item Sympathy and emotional support: Prayers, thoughts, and emotional support
\item Other useful information: Other useful information that helps understand the situation
\item Not related or irrelevant: Unrelated to the situation or irrelevant
\end{enumerate}

We settled on a blended prototype set of top-level actionable information types. Unlike CrisisNLP, this set imposes no specific ordering or prioritization on the data. This ordering is not applicable in all crises, nor indeed over time in the same crisis. For example, dead bodies are not a priority compared with the trapped and injured, on day zero; but on day 3-4, they may well be.

To avoid colouring user judgments, we took a conscious decision to make all tools as agnostic as possible, cutting out any biases that we noticed. This has the goal of placing all decisions about response and action on the analyst, instead of providing nudges about what has higher or lower rankings (as one implicitly does by ordering actionability types).

\begin{itemize}
\item A specific resource, where the kind of need is given explicitly.
\item Mentions of some group that's responding or aiding (e.g. volunteers, military, government, businesses, aid agencies).
\item Threats to the general crisis response. Weather warnings, fires, military action etc.
\item Changes in accessibility - limits to access, or other changes in how much transport can get through.
\item Damage to infrastructure, livelihoods etc.
\item Mention by name of the geographic areas affected
\item Changes in the local environment (weather, hazards, etc.); e.g., a storm is intensifying
\item General reporting about the rescue effort (from the media or the public)
\item Opinion or individual message.
\end{itemize}

Note that some categories, such as those around Needs, are collapsed into one. Following this example, we first should identify content about needs, and analysis of the exact type of information -- for example, the resource required, or the location -- can be postponed~(\cite{munro2011}).

\subsection{Gathering Actionability Data}

Labelling qualitative data with quantitative, hard categories is a difficult process, with some margin of error. The standard of this process directly reflects the quality of the resulting dataset and the performance of machine learning systems trained on them.

Humans typically perform rather poorly in data labelling when faced with more than seven labels, and best when faced with two~(\cite{sabou2014}). As we have nine categories, the set must be decomposed somehow. As with the informativeness task, we trialled various different combinations and setups using in-house volunteer annotators. We had best results from splitting types into two groups, and providing the ``personal opinion" option alongside both.

\begin{table}
\centering
\begin{tabular}{lc}
{\bf Category} & {\bf Count} \\
\hline
Needs & 22 \\
Response groups & 99 \\
Threats to response & 195 \\
Changes in accessibility & 19 \\
Damage to livelihoods & 62 \\
Mention of affected areas & 400 \\
Changes in environment & 40 \\
General reporting & 83 \\
Personal opinion & 423 \\
\end{tabular}
\caption{Actionability data}
\label{tab:actdata}
\end{table}

To maintain quality in crowdsourcing, a first portion of the data was annotated by experts who had experience in the goals of the data, crisis response, and linguistic annotation. After settling discrepancies, this pilot part of the data was converted to 118 ``gold" messages which would be used as test examples for crowd workers; for each gold example, the correct answers are pre-filled, and then used to give feedback on the job as crowd workers proceed. The test data gives crowd workers a safe environment in which to learn how to annotate, while seeing real examples of correctly-labelled data. This follows best practices in crowdsourced annotation of corpora~(\cite{sabou2014}).

We combined tweets from all days of Hurricane Irma\footnote{Kindly provided by PushShift -- http://pushshift.io} with data from CrisisLex, taking from the latter only crises of natural origin, in order to filter out short-term anthropogenic events like shootings. These were then filtered for informativeness based on the system introduced above, so that only informative tweets were scanned for actionable information. In total, 1350 tweets were labelled according to the actionability criteria indicated above, distributed as shown in Table~\ref{tab:actdata}.

Note that an individual tweet can contain more than one kind of information (or no relevant information at all). Therefore, the totals above do not sum to the dataset size. This comprises our dataset of actionable messages, annotated by a mixture of experts and crowd workers that have been trained with expert data.

\subsection{Extracting Actionable Content}
Having now a set of examples of actionable and non-actionable data, it is possible to train a system to triage messages according to actionability.

Deep learning requires large amounts of data; its advantage lies in a better ability to leverage better data. In this situation, however, we are faced with a paucity of examples from each actionability class, which is tough for deep learning. Indeed, applying a convolutional text kernel as with earlier informativeness analyses led to very poor results, with a F-score under 0.1 in almost every case. We attribute this to the extreme data sparsity.

Further, data where there is a strong imbalance is harder for most machine learning algorithms to learn to predict. In this case, our dataset of over 1000 instances often contains only a minority of positive examples, complemented by an overwhelming majority of negative examples. This is a difficult learning environment: for effective learning the number of positive and negative examples should be roughly equal.

We cast actionability extraction as a multi-class labeling problem. That is, a message can have more than one kind of actionable information at a time. For example,

{\em There are people camped out between Lincoln and fifth who need water}

contains both a need (type A) and a geographic location mentioned by name (type J) -- so it’s labeled [A,J]; hence, multi-class. We construct a solution based on many binary classifiers, one for each actionability type. Each classifier will attempt to predict whether or not the message contains a single type of actionability data, and a single classifier is dedicated to each kind of actionable information. Finally, the results are pooled to form an aggregate labeling of the message.

For feature extraction, following~\cite{aker2017}, we extracted from the data a set of keywords for each actionable information type, and also induced word embeddings over recent social media data as a continuous representation of individual word types. These were taken from 20 million tweets from 2016, filtered for English with langid.py~(\cite{lui2012}) randomly selected for variation in time (across seasons and across time of day), processed with GloVe~(\cite{pennington2014}), and having 25 dimensions (a low dimensionality is less prone to errors with this relatively sparse training data). Documents are converted to data by splitting them into words using a social media tokenizer~(\cite{oconnor2010}) and then, for each word in the keyword list, measuring the proportion of words in the document that are similar to the keyword. Similarity is defined by cosine vector proximity, with a cut-off of 0.45. That is to say, document words whose vectors have a cosine with the keyword under 0.45 are disregarded. Additionally, each document word is scaled by its score; so, if a ten-word document has one similar keyword, and the similarity is 0.6, the overall value for this keyword is 0.06 (10\% × 0.6). This feature extraction converts documents into a one-dimensional vector with a value for each keyword.

Sample keywords for two of the actionability types are:\footnote{ The full set is provided with the Emina toolkit}
\paragraph{Personal opinion:} {\tt [i people us all me please good will trump your pray praying toll concerned omg worried god]}
\paragraph{Changes in accessibility:} {\tt [G accessibility street bridge blocked derailment collapse close flooded closed careful cancel cancelled canceled avoid alert affect advised]}

These terms are those likely to be indicative of a particular class. They are extracted by finding the most discriminative words for each class through a simple Bayesian analysis. The full set of keywords and actionability types are downloadable from the Emina distribution and can be edited to suit new situations.

We needed a classifier that was rapid, could learn with with a low amount of data, and that did not select a liner decision boundary. To this end, the RBF SVM fit our needs. In addition, we reduced the ``negative" data (i.e. the non-actionable cases) down to match the number of positive examples for each class, randomly removing examples.

\subsection{Evaluation}

\begin{figure}
\centering
\includegraphics[width=0.45\columnwidth]{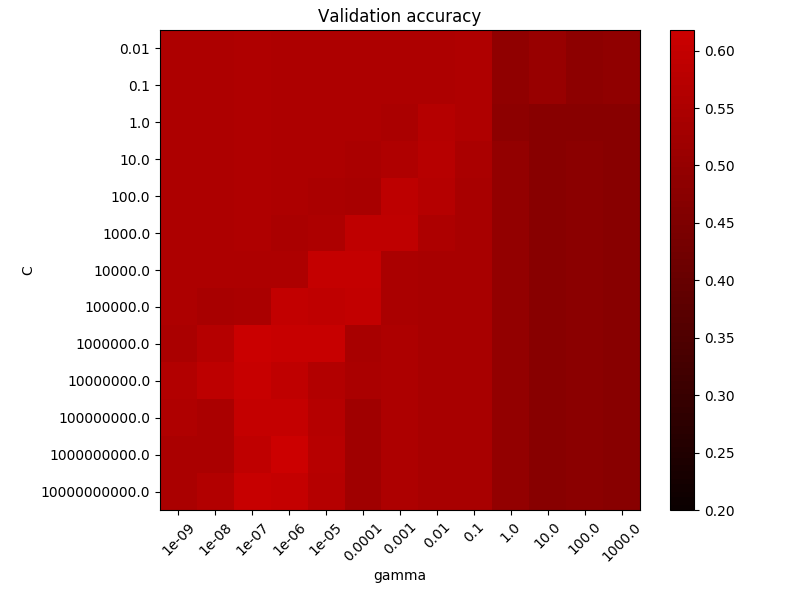}
\includegraphics[width=0.45\columnwidth]{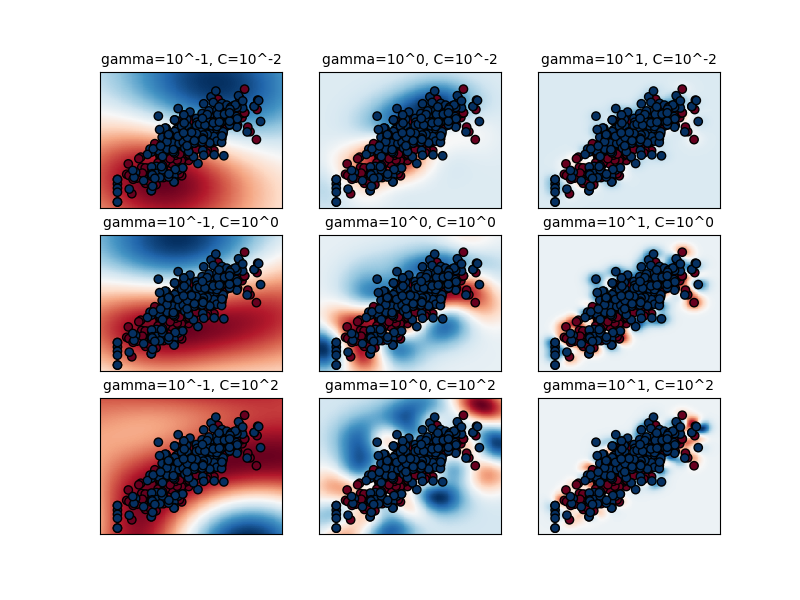}
\caption{Tuning parameters for the RBF SVM}
\label{fig:tuning}
\end{figure}

We measured both the accuracy of the system (i.e. the proportion of examples it got right) as well as its F1 score and its recall (i.e. the proportion of known actionable messages that were retrieved by the classifier). Recall was particularly important here as precise systems (i.e. those that return only actionable information) were not considered as useful as those with good recall; for us to correctly pick up two instances of people in need and ignore four hundred would still have high precision -- when no false need messages were returned (i.e. those where there is no need expressed) -- but poor impact on crisis analysis and response.

\begin{table}
\centering
\begin{tabular}{lccc|r}
{\bf Class} & {\bf Accuracy} & {\bf \underline{F1}} & {\bf Recall} & {\bf Baseline \underline{F1}}\\
\hline
Needs                      & 66.83\% & 57.50\% & 63.01\% & 45.21\% \\
Response groups            & 63.13\% & 62.94\% & 62.63\% & 40.18\% \\
Threats to response        & 57.18\% & 57.72\% & 58.46\% & 33.70\% \\
Change in accessiblity     & 57.89\% & 57.89\% & 57.89\% & 18.18\% \\
Damage to infrastructure, livelihoods & 56.45\% & 54.24\% & 51.61\% & 20.16\% \\
Geographic names           & 56.63\% & 57.84\% & 59.50\% & 45.23\% \\
Changes in environment     & 50.00\% & 47.37\% & 45.00\% & 23.58\% \\
Reporting about the rescue & 57.83\% & 58.82\% & 60.24\% & 13.95\% \\
Personal opinions          & 66.59\% & 63.43\% & 57.94\% & 42.27\%\\
\end{tabular}
\caption{Performance of actionability detectors}
\label{tab:act-acc}
\end{table}

Radial Basis Function (RBF) SVM had excellent performance in some categories and good performance in others. Other classifiers did best in some areas but much worse in others. RBF SVM’s overall performance indicates it is stable for this task especially given that we have relatively small datasets and the way messages are written varies very wildly across crises, even more so than to be expected in social media~(\cite{derczynski2013}).

The RBF SVM works by exploring the space out from each example, to see where boundaries lie. This effectively tells us the impact or sensitivity of each keyword, in the context of other keywords. It seems that such an approach is more effective than finding linear separators in such a data-imbalanced, high-variance environment.

RBF is tuned with two parameters, C and gamma. The gamma parameter defines the influence that a single data point has. C trades off error against simplicity -- a very complex system might make few errors, but may be fragile, having very poor performance on new data. The behavior of an RBF SVM can be highly sensitive to gamma; large values give individual examples power to distort the system’s behavior. To best tune our system, we searched for parameters (Figure~\ref{fig:tuning}. Results are for category G, access to infrastructure, though simplified for visualization to the first two keyword parameters.

The plots show that performance is best with a high C and a low gamma -- that is, we should not fit the data too tightly to either of these points. In the end, C=20 and gamma=3 worked for the general case. Overall results for RBF SVM are given in Table~\ref{tab:act-acc}. These are compared with a keyword-match baseline, which matches texts if any of the keywords for a given actionability type are present.

This actionability extraction, coupled with the informativeness filtering, is released as an open source tool Emina at https://github.com/GateNLP/emina.

\section{Profile of a Crisis}

Using both informativeness filtering and actionability classification, we are able to see how the kind of actionable information in a crisis situation develops over time. To this end, we applied informativeness filtering and then actionability extraction to messages around individual crises in prior datasets.

\subsection{Visualization}

\begin{figure}
\centering
\includegraphics[width=\columnwidth]{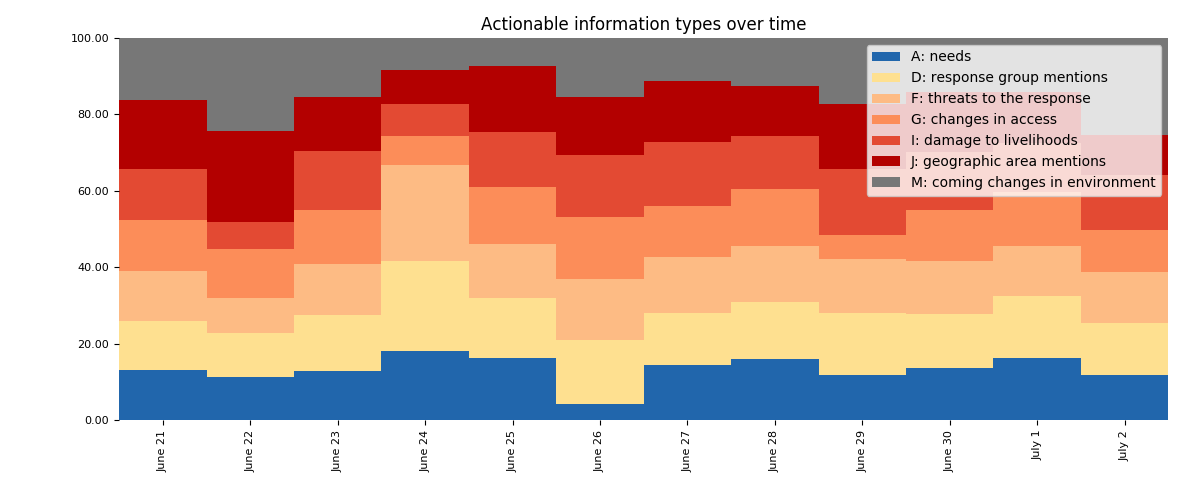}
\caption{Types of actionable data during the 2013 Alberta floods}
\label{fig:alberta}
\end{figure}

To measure the balance of actionable information over time, we bucketed messages according to their timestamps and then calculated the proportion of each actionability type. The results are shown in a proportional stacked bar chart. That is to say, for each time slot, the chart shows the proportion of actionable information during that period. If the dominant information type is ``needs", then this second will have a larger proportion than the others, regardless of volume. An example for the 2013 Alberta floods is shown in Figure~\ref{fig:alberta}. This kind of visualization is an accessible summary of the profile of the crisis over time.

\section{Conclusion}

Filtering out uninformative messages is crucial to effective use of human monitors of electronic messages during a crisis. The majority of inputs in recent crises are irrelevant, and systems that preprocess and reduce the volume of irrelevant messages become invaluable. We developed a pair of techniques that first remove the bulk of irrelevant messages, and then attempt to select particular kinds of actionable messages, according to some top-level actionability categories.

We were able to extract informative messages at over 90\% accuracy, incorporating both the accurate selection of informative content and also avoiding extraction of non-informative content. Furthermore, we were able to recall between 50\% and 70\% of actionable information over eight individual information classes.
The results presented in this paper are a first step to examine the feasibility of automatically identifying informative, and actionable messages such as those present from social media sources during a crisis event. While the results in this paper are promising, further research is needed to improve this approach. More importantly, given the importance of data and responsibilities of crisis responders, additional research is needed into the risks and ethics associated with automatic data processing.  With automated services and tools, like Emina presented in this paper, we gain the potential to support crisis responders, digital volunteers and communities themselves in processing large volumes of data. 

\section*{Acknowledgments}
This work was supported by funding from the European Commission through grant agreement 687847, {\sc Comrades}. The authors are grateful to CrowdFlower and Rob Munro for their continued support and the resources generously donated which enabled creation of these datasets, as well as to Pushshift.io and Jason Baumgartner, who helped provide the Irma data rapidly during and after the crisis.

\printbibliography

\end{document}